\documentclass{article}
\usepackage{ijcai25}

\usepackage[symbol]{footmisc}
\usepackage{times}
\usepackage{soul}
\usepackage{url}
\usepackage[hidelinks]{hyperref}
\usepackage[utf8]{inputenc}
\usepackage[small]{caption}
\usepackage{graphicx}
\usepackage{amsmath}
\usepackage{amsthm}
\usepackage{booktabs}
\usepackage{algorithm}
\usepackage{algorithmic}
\usepackage[switch]{lineno}
\usepackage{subfig}

\usepackage{url}
\usepackage{color}
\usepackage{xcolor}
\usepackage{pifont}% http://ctan.org/pkg/pifont
\usepackage{comment}
\usepackage{amsfonts}

\def\ie{{i.e.}}

\renewcommand{\paragraph}[1]{{\noindent \bf {#1}.~}}

\newcommand{\ym}[1]{\textcolor{blue}{#1}}
\newcommand{\name}{{\textsc{KDC}}}

\title{Keypoints as Dynamic Centroids for Unified Human Pose and Segmentation}

\author{
Niaz Ahmad$^1$\and
Jawad Khan$^2$\and
Kang G. Shin$^{3}$\and
Youngmoon Lee$^{4^*}$ \And
Guanghui Wang$^1$\\
\affiliations
$^1$Toronto Metropolitan University, 
$^2$Gachon University,
$^3$University of Michigan,
$^4$Hanyang University\\
\emails
\{niazahmad89, wangcs\}@torontomu.ca,
jkhanbk1@gachon.ac.kr,
kgshin@umich.edu,
youngmoonlee@hanyang.ac.kr
}
\begin{document}

\maketitle

\footnotetext[1]{Youngmoon Lee is the corresponding author}

\begin{abstract}

The dynamic movement of the human body presents a fundamental challenge for human pose estimation and body segmentation. State-of-the-art approaches primarily rely on combining keypoint heatmaps with segmentation masks but often struggle in scenarios involving overlapping joints during pose estimation or rapidly changing poses for instance-level segmentation. 
To address these limitations, we leverage {\em Keypoints as Dynamic Centroid} ({\name}), a new centroid-based representation for unified human pose estimation and instance-level segmentation. 
{\name} adopts a bottom-up paradigm to generate keypoint heatmaps for easily distinguishable and complex keypoints, and improves keypoint detection and confidence scores by introducing KeyCentroids using a keypoint disk. It leverages high-confidence keypoints as dynamic centroids in the embedding space to generate MaskCentroids, allowing for the swift clustering of pixels to specific human instances during rapid changes in human body movements in a live environment. Our experimental evaluations focus on crowded and occluded cases using the CrowdPose, OCHuman, and COCO benchmarks, demonstrating {\name}'s effectiveness and generalizability in challenging scenarios in terms of both accuracy and runtime performance. Our implementation is available at
\end{abstract}

\section{Introduction}
\label{sec:intro}
Human pose estimation and body segmentation are crucial for human-computer interaction and real-time visual human analysis. The primary objective is to identify individuals and their activities from 2D joint positions and body shapes. The underlying main challenges include handling an unknown number of overlapping, occluded, or entangled individuals and managing the rapidly increasing computational complexity as the number of individuals grows \cite{han2025occluded}. Human-to-human interactions further complicate spatial associations due to limb contact and obstructions, necessitating an efficient, scalable, and accurate unified model for human pose and segmentation.

In this paper, we propose {\name}, %Keypoints as Dynamic Centroid, 
a new centroid-based unified representation for human pose 
estimation and instance-level segmentation. 
It first detects individual keypoints in a bottom-up manner and then 
employs high-confidence keypoints as dynamic centroids for mask 
pixels to perform instance-level segmentation.
%It uses PoseNet to detect individual keypoints in a bottom-up manner.
%Moreover, SegNet is designed to use the high confident 
%CCCCC Why SeNet? Do you mean \name?
%keypoint as a dynamic centroid for mask pixels.   
% bottom-up pose estimator to detect individual keypoints directly by employing the PoseNet, and perform pixel-level classification by employing the SegNet where the detected keypoints are used as a center of attraction to associate the pixels to the right instance.
% \textcolor{green}{If possible make short the above para ?}
Unlike top-down approaches \cite{ref35,ref36,ref94}, 
KDC detects humans without requiring a box detector 
or incurring runtime complexity. 
%%%%%%%%%%%%%%%%%%%%%%%%%%%%%%%%%%%%%%%%%%%%%%%%%%%%%%%%%%%%%%%%%%%%%%%%%%%%
%, enabling effective pose estimation and instance-level segmentation.
%This approach detects the human body \emph{without} a bounding box concept, thus enabling effective pose estimation, along with instance segmentation, without incurring the runtime complexity of the top-down approach.
%%%%%%%%%%%%%%%%%%%%%%%%%%%%%%%%%%%%%%%%%%%%%%%%%%%%%%%%%%%%%%%%%%%%%%%%%%%%%%%%%%%%%%%%%%%%%%%%%%%%%

%%%%%%%%%%%%%%%%%%%%%%%%%%%%%%%%%%%%%%%%%%%
\begin{figure}[t!]
%\begin{minipage}{1.5\columnwidth}
\centering
\includegraphics[width=7.8cm,height=5.5cm]{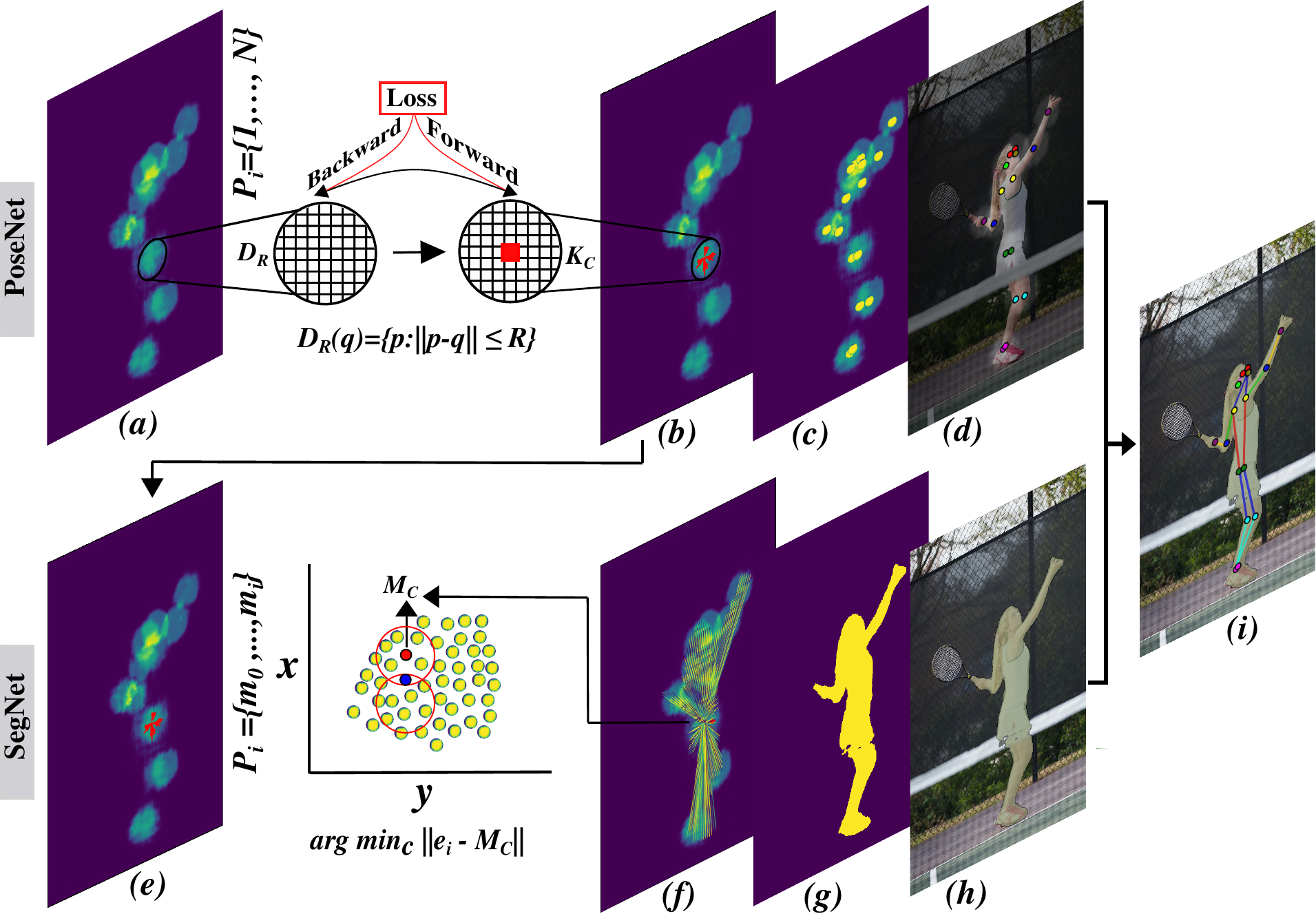}

\caption{
PoseNet operation begins by generating keypoint heatmaps in the feature space using a disk representation \( D_R \) to identify potential keypoint locations. It then introduces KeyCentroid to refine these keypoint coordinates to improve accuracy. 
SegNet leverages the KeyCentroid \( K_c \) defined by PoseNet to establish MaskCentroid \( M_c \), which is essential for clustering mask pixels corresponding to specific human instances.}

%  \caption{PoseNet generates (a) keypoint heatmaps in 
%  feature space using keypoint disk $D_R$, introduces 
%  (b) KeyCentroid $K_c$ for each feature map, and 
%   backpropagates across the network to the position 
%   $D_R$ to obtain the (c) optimal keypoint coordinate 
%   for (d) human pose estimation. 
%   SegNet uses KeyCentroid to (e) define MaskCentroid $M_c$ 
%   that helps
%  (f) cluster the mask pixels to a particular human instance, 
%  (g) establish a high-level feature map of semantic 
%  segmentation, and generate instance-level segmentation. 
%  The unified structure of (i) human pose and instance-level 
%  segmentation is obtained from the high-level features of 
%  PoseNet and SegNet.
% }
\label{fig:PoseSegNet}
\end{figure}

KDC is not the first to leverage bottom-up approach 
\cite{ref52,ref86,ref34,ref64,2025visualcent}.
However, the model in \cite{ref52} employs human poses to refine 
pixel-wise clustering for segmentation, and thus 
does not perform segmentation well in segmentation tasks. 
Other models suffer from the computational overhead of a 
person detector \cite{ref34}, the scalability problem 
for instance-level segmentation \cite{ref64}, and model 
complexity \cite{ref86}, making them unsuitable for crowded 
scenarios and real-time applications.
%YM: We need clear comment about ours over top-down and bottom-up; check my attempt below. I agree with Professor. 
%Unlike existing solutions, {\name} does not incur high overheads of top-down approach due to person detector, and segmentation performance and scalability issues of bottom-up approach due to pixel-wise clustering.
%
Unlike these models, KDC avoids the computational overhead of a 
person detector and suffers from neither the degraded segmentation 
performance nor the scalability problem of pixel-wise clustering.

KDC overcomes these problems using two primary networks: {\em PoseNet}, 
which generates keypoints, and {\em SegNet}, which produces segmentation 
masks using high-confidence keypoints (Fig. \ref{fig:PoseSegNet}). 
PoseNet creates keypoint heatmaps using a keypoint disk representation 
that estimates the relative displacement between pairs of keypoints, 
enhancing the precision of long-range, occluded, and proximate keypoints 
(Fig.~\ref{fig:PoseSegNet}a). A KeyCentroid is defined for each keypoint 
heatmap locus offset vectors to the centroid of each keypoint disk, 
helping KDC identify the precise human keypoint coordinates 
(Fig.~\ref{fig:PoseSegNet}b). Additionally, KDC calculates the 
keypoint confidence score using the precise keypoint coordinates 
(Fig.~\ref{fig:PoseSegNet}c), with the final predicted keypoints illustrated in (Fig.~\ref{fig:PoseSegNet}d).

Meanwhile, SegNet performs pixel-level classification using 
dynamic high-confident keypoints as MaskCentroids 
(Fig.~\ref{fig:PoseSegNet}e). MaskCentroid defines an embedding space 
that associates pixels with the correct instance 
(Fig.~\ref{fig:PoseSegNet}f) and generates high-level semantic maps 
(Fig.~\ref{fig:PoseSegNet}g). Leveraging these semantic maps, the 
system produces instance-level segmentation (Fig.~\ref{fig:PoseSegNet}h). 
The PoseSeg module combines high-level features from both PoseNet and 
SegNet to provide a unified representation of human pose and 
instance-level segmentation (Fig.~\ref{fig:PoseSegNet}i).

We evaluated the performance of the KDC using the CrowdPose \cite{ref92}, 
OCHuman \cite{ref64}, and COCO \cite{ref2} benchmarks. 
To the best of our knowledge, KDC is the first 
real-time model with reliable performance to offer a
unified representation of human pose estimation and 
instance-level segmentation. 
%To the best of our knowledge, {\name} is the first high run-time and reliable application for the task of human pose estimation and instance segmentation. 
% Our multitask system outperforms existing pose and segmentation models using the COCO dataset, achieving $0.731$ mean average precision (mAP) for human pose and $0.456$ mAP for instance segmentation. While using the OCHuman dataset \cite{ref64} with heavily occluded humans, {\name} achieves $0.430$ mAP for human pose and $0.556$ mAP for instance segmentation.
%}
%\textcolor{red}{
%{\name} surpasses the best prior bottom-up technique \cite{ref3} by increasing the average precision from $0.696$ to $0.725$.}
%\ym{YM: try to organize the contribution match with the overall paper organization (sections/subsections) of the paper such that readers can easily understand the relevant sections where you make specific contribution.}
%////////////////////////////////////////////
%\ym{YM:You can reorganize the contribution and organization of the main body but it should be clearly connected step-by-step, section-by-section.}
%
%\paragraph{Contribution.} To the best of our knowledge, {\name} is the robust to perform the task of joint pose estimation and instance segmentation.
%In summary,
This paper makes the following contributions.
\begin{itemize}
    %\item We introduce the strong keypoint heat map to detect both soft and hard keypoints to accurately estimate human pose, and the body heat map for locating the individuals and increasing the keypoint confidence scoring; % (Section \ref{sec:PosePipeline});
	%\item We investigate the effects of various factors that contribute to single and multi-person pose estimation in bottom-up approach (Section \ref{subsec:Bottom-Up});
	%
	%\item We propose a novel and efficient {\name} approach that generates SKHM for both soft and hard keypoints to accurately estimate human pose (Section \ref{subsec:SKHM});
	%
	%\item {\name} generates BHM as the same manner as SKHM in order to localize the individuals in the image (Section \ref{subsec:BHM});
	%
		
\item %\red {
   The development of KeyCentroid, a novel method that directs keypoint 
   vectors towards the centroid within the keypoint disk. This approach 
   helps identify the precise keypoint coordinates in human pose 
   estimation, thereby enhancing confidence in the results (\S{3.2}).

   % Development of KeyCentroid, a novel approach that pushes 
   % keypoint vectors to the centroid in the keypoint disk. 
   % This approach helps pinpoint precise keypoint 
   % coordinates in human pose estimation, 
   % enhancing confidence in the results 
   % (\S{3.2});
    
    %which helps to identify the precise keypoint coordinates with a high confidence for human pose estimation 
    
	 % KeyCentroid defines 2D offset vectors points to the centroid of each keypoint feature map, which helps to identify the precise keypoint coordinates for human pose estimation (\S\ref{subsec:KeyCentroid});
	%pointing at the center of attraction for each keypoint.  Thus helps to identify the precise human keypoint coordinates;
	%}
	
\item 
   The developement of MaskCentroid leverages high-confidence keypoints 
   as dynamic centroids for mask vectors in the embedding space. 
   This approach effectively associates pixels with the correct instance, 
   even during rapid changes in human body movements (\S{3.3}).

	% Development of MaskCentroid, leveraging highly 
 %        confident keypoint as a dynamic centroid for the mask vectors in the embedding space, that helps to effectively associate 
 %        pixels with the right instance due to the rapid changes in human body actions
 %     %in case of rapid changes in the human body 
 %     (\S{3.3});

	%}
	%\item {\name} introduces a MaskCentroid to define an embedding representation for the pixel associations to identify the human body shape structure;% (Section \ref{sec:SegPipeline});
	%
	%\item We utilize a refine network to produce a refine keypoints and instance segmentation mask, and introduce a new pose and instance segmentation algorithm to visualize the joint human pose and instance segmentation;% (Section \ref{sec:ISP});
	%
	%\item We define dense logistic network (DLN) as a refinery network to minimizes the loss between the predicted (keypoint, mask) with the ground truth to produce a refine keypoint and instance segmentation mask (Section \ref{subsec:DLN});
	%
	%\item We introduce a novel algorithm, Instance Segmented Pose (ISP), to present human estimated pose along with instance segmentation mask \ym{using both pipelines} (Section \ref{subsec:ISP});
	%
\item %\red {
   An in-depth evaluation (\S{4}) and ablation studies (\S{5}) 
   demonstrate the effectiveness of the unified representation of 
   human pose and instance-level segmentation.
	%}
\end{itemize}

 % \textcolor{green}{The Contributions need to refine more if possible, like what actually we want to propose/ (write the name of designed/proposed model}
%%%%%%%%%%%%%%%%%%%%%%%%%%%%%%%%%%%%%%%%%%%%%%
%%%%%%%%%%%%%%%%%%%%%%%%%%%%%%%%%%%%%%%%
\begin{figure*}[t!]
%\begin{minipage}{1.5\columnwidth}
\centering
  \includegraphics[width=16cm,height=5.5cm]{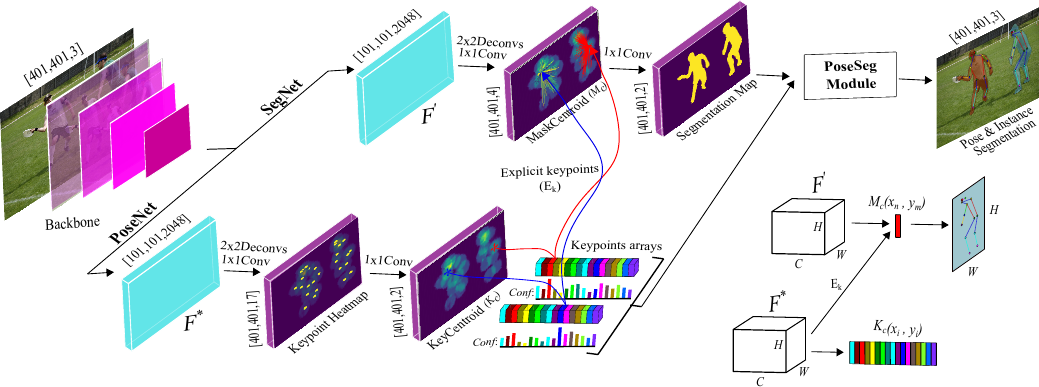}

\caption{The overview of the proposed KDC model. PoseNet generates 
keypoint heatmaps and refines them with KeyCentroid \( K_c \), 
improving keypoint accuracy. SegNet uses \( K_c \) to create MaskCentroid 
\( M_c \), clustering mask pixels for precise instance segmentation.
The PoseSeg module integrates these outputs, resulting in accurate 
unified human pose estimation and instance-level segmentation.}

% \caption{
% The overview of the proposed KDC model. It uses CNN backbone to extract pose and segmentation features. PoseNet generates keypoint heatmap  and KeyCentroid $K_c$ using keypoint disk to 
% predict the optimal 2D keypoint coordinates and enrich the 
% confidence score. SegNet produces a segmentation map using 
% the dynamic explicit keypoints as a MaskCentroid $M_c$ for 
% the pixels in the embedding space to assign the pixel to the 
% right instance. Finally, the PoseSeg Module uses the 
% features from both networks for the unified representation of human pose and instance-level segmentation.}
%  %, i.e., human pose and instance segmentation.
        
	\label{fig:overview}
\end{figure*}
%%%%%%%%%%%%%%%%%%%%%%%%%%%%%%%%%%%%%%%%%%%%%%%%%%%%%%%%%%%%
% \begin{figure}[t!]
% %\begin{minipage}{1.5\columnwidth}
% 	\centering
% 	\includegraphics[width=7cm,height=5cm]{figures/Net3_exp.pdf}
% 	\caption{(a) Indicates MaskCentroid a dynamic high confident keypoint (b) presents a precise segmentation map (c) indicates instance-level segmentation and (d) shows unified representation of human pose and estimation.}
%  %, i.e., human pose and instance segmentation.
        
% 	\label{fig:D_Mask_Centroid}
% \end{figure}
%%%%%%%%%%%%%%%%%%%%%%%%%%%%%%%%%%%%%%%%%
\section{Related Work}
\label{sec:related}

%\ym{YM:Related work is too long for 8-page submission.}

%The recent literature in the field of sentiment analysis can be broadly divided into supervised algorithms, semi-supervised or un- supervised methodologies, active learning and domain adaptation approaches

\paragraph{Human Pose Estimation}
Approaches for human pose estimation can be 
classified as top-down or bottom-up.
The top-down approach first runs a human detector and then 
identifies keypoints. Representative works include HRNet 
\cite{ref76}, RMPE \cite{ref36}, Multiposenet \cite{ref51}, convolutional pose machine 
\cite{ref31}, CPN \cite{ref35}, Mask r-cnn \cite{ref34}, 
simple baseline \cite{ref70},  CSM-SCARB \cite{ref80}, 
RSN \cite{ref81}, and Graph-PCNN \cite{ref82}. 
The top-down approach explores the human pose in a 
person detector, thus achieving a satisfactory 
performance, but it is computationally expensive.
The bottom-up approach like DeepCut \cite{ref27} and 
DeeperCut \cite{ref28}, unlike the top-down counterpart, 
detects the keypoints in a one-shot manner. 
It formulates the association between keypoints as 
an integer linear scheme which takes a longer processing 
time. Part-affinity field techniques like OpenPose 
\cite{ref32} and other extensions, such as PersonLab 
\cite{ref52}, and HGG \cite{ref74} have been developed 
based on grouping techniques that often fail in crowd. 
{\name} aims to specifically improve hard 
keypoint detection in crowded and occluded cases 
%and generally soft keypoint 
by introducing the keypoint heatmaps using keypoint 
disks and KeyCentroid. %, in our pose estimation pipeline. }

\paragraph{Instance-level Segmentation}   
Instance-level segmentation is done in either single-stage 
\cite{ref57,ref58,ref59} or multi-stage \cite{ref34,ref39}. 
The single-stage approach generates intermediate and 
distributed feature maps based on the input image.
InstanceFCN \cite{ref57} produces instance-sensitive scoring 
maps and applies the assembly module to the output instance. 
This approach is based on repooling and other non-trivial 
computations (e.g., mask voting), which is not suitable 
for real-time processing. 
YOLACT \cite{ref59} runs a set of mask prototypes and 
uses coefficient masks, but this method is critical 
to obtain a high-resolution output. 
The multi-stage approach follows the detect-then-segment 
paradigm. It first performs box detection, and then pixels 
are classified to obtain the final mask in the box region. 
Mask R-CNN \cite{ref34} is based on multi-stage instance 
segmentation that extends Faster R-CNN \cite{ref39} by 
adding a branch for predicting segmentation masks for 
each Region of Interest. 
The subsequent work in \cite{ref68} improves the accuracy 
of Mask R-CNN by enriching the Feature Pyramid Network 
\cite{lin2017feature}. 
In contrast, our segmentation pipeline introduces 
MaskCentroid, a dynamic clustering point that helps  
cluster the mask pixels to a particular instance under 
the rapid changes in human-body movements.  

\paragraph{Joint Human Pose and Instance-level Segmentation}  
%
%\textcolor{red}{ Researchers have also worked on the joint pose estimation and instance segmentation problem []. 
%State-of-the-art developments have been made in human pose estimation and instance-level segmentation. 
In the line of multi-task learning paradigm, joint pose 
estimation and instance-level segmentation have 
received significant attention in recent years.
Mask R-CNN \cite{ref34} was the first pioneer method,
but it suffers from high computational costs due to 
its top-down nature.
% Pose2Seg \cite{ref64} proposes human pose-based instance-level segmentation by separating instances based on the human pose, rather than the proposal region; however, it takes already generated pose as input that makes concerns in end-to-end training model.
%this is not joint model, we can drop if we need space.
%It takes already generated pose as input that makes concerns in end-to-end training model. 
%This is important
%
PersonLab \cite{ref52} and Pose\textit{Plus}Seg 
\cite{ref86} are closest to {\name}. 
Both of them can be considered as end-to-end joint pose and 
instance-level segmentation models that use a bottom-up approach.
However, there are several major differences that make {\name} 
more effective, scalable, and real-time.
First, they rely on static features to detect or group keypoints by 
using greedy decoding; in contrast, {\name} introduces KeyCentroid that 
calculates the optimal keypoint coordinates, and uses MaskCentroid, 
a dynamic clustering point for instance-level segmentation.
Second, their segmentation does not perform well on highly entangled 
instances due to part-induced geometric embedding descriptors. 
% for human class instance-level segmentation. 
Finally, they involve the complex structure model with a couple of 
refined networks, making them infeasible for real-time purposes.
%We propose a simple, yet effective system to handle the above complications by introducing KeyCentroid that calculates the optimal keypoint coordinates using keypoint disk, and MaskCentroid, a dynamic clustering point for instance-level segmentation. 

%PersonLab \cite{ref52} groups keypoints by using greedy decoding. 
%This method also reports a part-induced geometric embedding descriptor for human class instance-level segmentation. 
%However, this approach fails to perform segmentation on highly entangled instances. 
%This is important
%Pose\textit{Plus}Seg \cite{ref86}played an important role in this regard; however, it compromises the performance of the model using static centroid and a couple of refined networks making it a complex structure model.
%
%We propose a simple, yet effective system to handle the above complications by introducing KeyCentroid that calculates the optimal keypoint coordinates using keypoint disk, and MaskCentroid, a dynamic clustering point for instance-level segmentation. 

% \input{overview}

\section{Technical Approach}
\label{sec:method}

\subsection{Keypoint Heatmap using Disk Representation}
\label{subsec:SKHM}

KDC generates keypoint heatmaps using disk representation (KHDR) through 
PoseNet, forming the foundation for human pose estimation (Fig.~3a). 
In this phase, individual keypoints are detected and aggregated in the 
output feature maps. We adopt a residual-based network for a 
multi-person pose setting to produce keypoint heatmaps—one channel per 
keypoint—and KeyCentroid, with two channels per keypoint for vertical 
and horizontal displacement within the keypoint disk.

The keypoint prediction approach is as follows: Let \( p_i \) represent 
the keypoint position in the image, where \( i \in \{1, \dots, N \} \) 
corresponds to the 2D positions of the pixels. 
A keypoint disk \( D_R(q) = \{ p : \| p - q \| \leq R \} \) of 
radius \( R \) is focused at point \( q \), centered in the disk. 
Similarly, \( q_{j,k} \) signifies the 2D position of the \( j \)th 
keypoint of the \( k \)th person instance, where \( j \in 
\{1, \dots, I \} \) and \( I \) is the number of individual keypoints 
in the image. A binary classification approach is followed for each known 
keypoint \( j \). Specifically, every predicted keypoint pixel \( p_i \) 
is binary classified such that \( p_i = 1 \) if \( p_i \in D_R \) for 
each person keypoint \( j \); otherwise, \( p_i = 0 \). Independent dense 
binary classification tasks are performed for each keypoint, leading to 
distinct keypoint maps.

During the training process, the heatmap loss is computed using the 
binary cross-entropy (logistic loss) function defined as:

\begin{equation}
\mathcal{L}_{\text{heatmap}} = - \frac{1}{N} \sum_{i=1}^{N} 
\left[ y_i \log(\hat{y}_i) + (1 - y_i) \log(1 - \hat{y}_i) \right],
\end{equation}
where \( N \) is the total number of pixels, \( y_i \) is the true 
binary label for pixel \( p_i \), and \( \hat{y}_i \) is the predicted 
probability that pixel \( p_i \) belongs to the keypoint. 
This loss function measures the difference between the predicted 
probability and the true label, and the average loss across all pixels 
in the heatmap is used to train the model. Back-propagation is performed 
throughout the entire image, except for regions that encompass individuals 
lacking comprehensive keypoint annotations (e.g., crowded and small-scale 
person segments).

\begin{figure}[t!]
    \centering
	\captionsetup[subfigure]{justification=centering}
	\subfloat[Keypoint heatmap]{
		\includegraphics[width=2.6cm, height=2.7cm ]{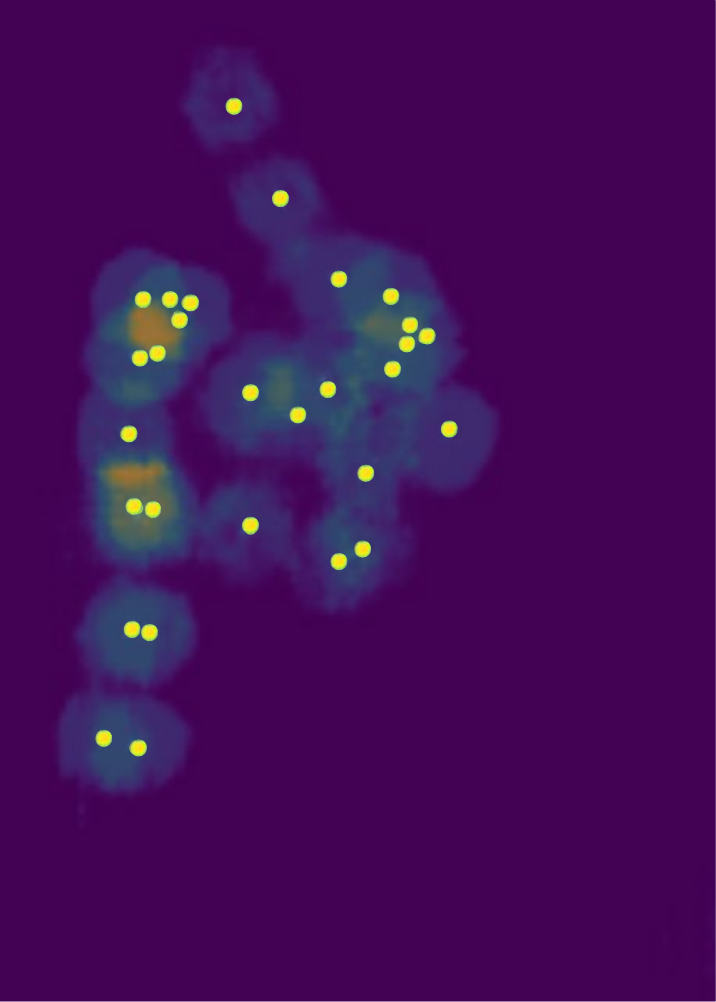}
		\label{fig:input_imag}}
	\subfloat[PGO]{
		\includegraphics[width=2.6cm, height=2.7cm]{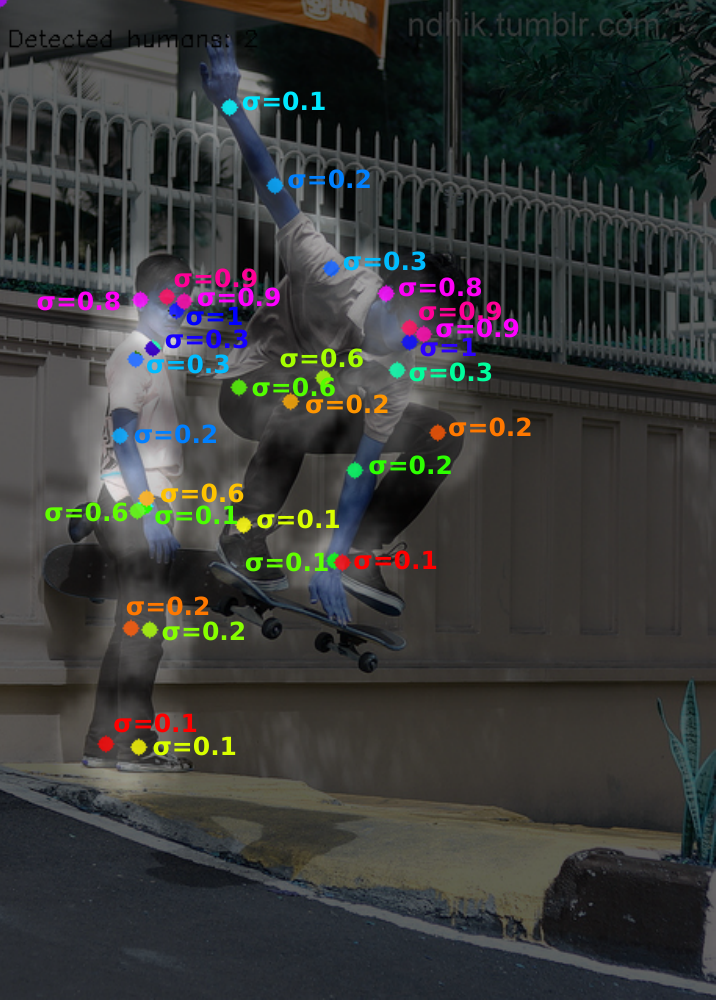}
		\label{fig:point_wise}
    }
        \subfloat[KeyCentroid]{
		\includegraphics[width=2.6cm, height=2.7cm]{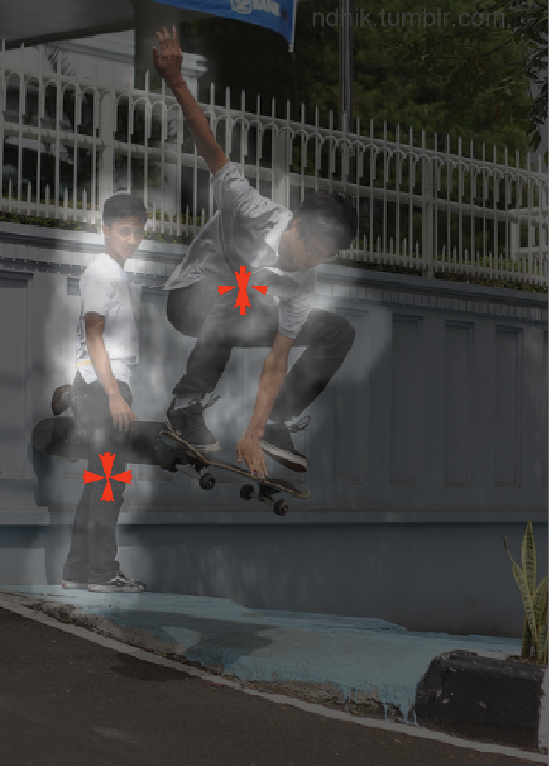}
		\label{fig:KeyCentroid2}
    }
\caption{(a) presents Keypoint heatmap using keypoint disk, 
    (b) shows Point-wise Gaussian optimization (PGO) where $\sigma$ 
    values are defined for each keypoint (c) Indicates KeyCentroid 
    defined for the right knee using the keypoint disk.}
	\label{fig:3}
\end{figure}

\paragraph{Point-wise Gaussian Optimization}
\label{subsec:PGO}
To obtain optimal keypoint coordinates, we apply a Gaussian smoothing 
technique \cite{ref84} for each individual keypoint, referred to as 
\emph{point-wise Gaussian optimization}. This approach effectively 
reduces noise while preserving valuable information, producing the 
keypoint heatmap as:

\begin{equation}
G(x,y) = \dfrac{1}{2\pi{\sigma}^2} e^{-\dfrac{x^2+y^2}{2{\sigma}^2}},
\end{equation}
where \( G(x,y) \) is the Gaussian kernel, \( \sigma \) is the standard 
deviation of the distribution, and \( x \) and \( y \) represent the 
2D keypoint coordinates. We define the \( \sigma \) range from 0.1 to 1 
to accommodate variations among keypoints. For high-variance keypoints 
(HVK) such as the wrist, ankle, elbow, and knee, we set \( 0.1 \leq 
\sigma < 0.5 \). Conversely, for low-variance keypoints (LVK) like the 
nose, shoulder, and hip, we set \( 0.5 \leq \sigma < 1 \), as depicted 
in Fig.~\ref{fig:point_wise}.

A smaller \( \sigma \) value, close to 0.1, intensifies pixel values 
of keypoints, proving effective in congested and intricate scenarios. 
In contrast, a larger \( \sigma \) value, close to 1, yields optimal 
results in less crowded cases.
Our analysis investigates how the \( \sigma \) value impacts 
system performance in ablation studies (\S{5.3}).

%%%%%%%%%%%%%%%%%%%%%%%%%%%
%%%%%%%%%%%%%%%%%%%%%%%%%%%

\subsection{KeyCentroid}
\label{KeyCentroid}
In addition to keypoint heatmaps, our PoseNet, in conjunction with the 
residual network, introduces KeyCentroid \( k_{c} \) for each keypoint 
as shown in Fig.~\ref{fig:overview}. The objective of KeyCentroid 
is to improve both the accuracy of keypoint localization and the 
confidence scores.

For each keypoint pixel \( p_i \) within the disk \( D_R \), the 2D 
KeyCentroid vector \( k_{v} = q_{j,k} - p_i \) originates from the 
pixel position \( p_i \) and points to the \( j^{th} \) keypoint 
of the \( k^{th} \) person instance, as illustrated in 
Fig.~\ref{fig:KeyCentroid2}.
We generate a vector field within \( D_R \) by solving a 2D 
regression problem for the \( j^{th} \) keypoint with spatial 
coordinates \( (x_j, y_j) \), and compute its response on the 
ground truth feature map \( F^{*}_{j} \) as:

\begin{equation}
F^{*}_{j}(x, y) = \exp\left(-\dfrac{(x - x_j)^2 + (y - y_j)^2}{2\sigma^2}\right),
\end{equation}
where \( \sigma^2 \) is the variance related to the disk radius 
\( R = 32 \), used to normalize the KeyCentroid and align its 
dynamic range with the keypoint heatmap loss.

During training, we penalize the KeyCentroid error using the
L1 loss function, which is defined as:
\begin{equation}
\mathcal{L}_{\text{KeyCentroid}} = \frac{1}{N} \sum_{i=1}^{N} \| k_{v,i} - \hat{k}_{v,i} \|_1,
\end{equation}
where \( N \) is the number of pixels in the disk \( D_R \), 
\( k_{v,i} \) is the ground truth KeyCentroid vector for pixel 
\( p_i \), and \( \hat{k}_{v,i} \) is the predicted KeyCentroid 
vector. This loss function measures the difference between the 
predicted and true KeyCentroid vectors, and the average loss 
across all pixels in the disk is used to train the model.

The error is back-propagated for each pixel \( p_i \in D_R \).
We then aggregate the keypoint heatmap and KeyCentroid to determine 
the optimal keypoint coordinates \( (x_j, y_j) \), which improves the 
detection of both easily distinguishable and challenging keypoints.
Our ablation experiments examine the impact of our uniquely designed 
KHDR and KeyCentroid on keypoint detection (\S{5.1}).

{\subsection{MaskCentroid} %Centroid
\label{subsec:MaskCentroid}

Instance-level segmentation is a pixel classification problem focused on allocating pixels to the correct instance. We introduce MaskCentroid \( C_i \) (a dynamic high-confidence keypoint), as illustrated in Fig. 4a. Our mechanism clusters mask pixels using the defined centroid \( C_i \) inside each annotated person, pointing from the image position \( x_i \) to the centroid \( C_i \) of the corresponding instance. At each semantically identified human instance, the pixel embedding \( e(x_i) \) reflects a local approximation of each mask pixel’s absolute location relative to the individual it pertains to, effectively capturing the anticipated 2D structure of the human body. 

Consequently, for every pixel, we determine pixel offsets pointing to \( C_i \). Each \( C_i \) serves as a high-confidence keypoint that can change with the rapid variation in keypoints, as shown in Fig. 4a. The objective of human-body segmentation is to assign a set of pixels \( P_i = \{m_0, m_1, m_2, \dots, m_i\} \) and its 2D embedding vectors \( e(m_i) \) into a set of instances \( I = \{N_0, N_1, N_2, \dots, N_j \} \) to generate a 2D mask for each human instance, as shown in Fig. 4b. Pixels are clustered to their corresponding centroid \( C_i = \frac{1}{N} \sum_{m_i \in N_j} m_i \).
This is achieved by defining a pixel offset vector \( v_i \) for each known pixel \( m_i \), so that the resulting embedding \( e_i = m_i + v_i \) points from its respective instance centroid. We penalize pixel offset loss using the L1 loss function during model training:

\begin{equation}
\mathcal{L}_{\text{offset}} = \frac{1}{N} \sum_{i=1}^{N} \| e_i - (m_i + v_i) \|_1.
\end{equation}

To cluster the pixels to their centroid, it is important to specify the positions of the instance centroids and assign pixels to a particular instance centroid. We employ a Gaussian function \( \phi_j(e_i) \) for each instance \( N_j \), which converts the distance between a pixel embedding \( e_i = m_i + v_i \) and the instance centroid \( C_i \) into a probability of belonging to that instance:

\begin{equation}
\phi_j(e_i) = \exp \left( - \frac{\| e_i - C_i \|^2}{2 \sigma_j^2} \right).
\end{equation}

% A higher probability indicates that the pixel embedding 
% $e_i$ is in proximity to the instance centroid, and is 
% likely to belong to that instance. In contrast, a low 
% probability means that the pixel is more likely to belong 
% to the background (or another instance). Specifically, if 
% $\phi_{j}(e_i) > 0.5$, then that pixel, at location $x_i$, 
% will be assigned to instance $N_j$.

% A high probability means that the pixel embedding $e_i$
% is close to the instance centroid and is likely to belong to that instance, while a low probability means that the pixel is more likely to belong to the background (or another instance). More specifically, if $\phi_{k}(e_i) > 0.5$, than that pixel, at location $x_i$, will be assigned to instance $k$.
% We use a density-based clustering algorithm \cite{ref83} to first locate a set of centroids as a center of attraction. Having obtained an array of centroids $\mathcal{C} = \{C_0, C_1,..., C_i\}$, we add pixels to a particular instance based on a minimum distance-to-centroid:
%
% \begin{equation}
% e_i \in m_i : i = \mbox { arg min}_{\mathcal{C}}  \|e_i-C\|.
% \label{eq:4}
% \end{equation}
%%%%%%%%%%%%%%%%%%%%%%%%%%%%%%%%%%%%%%%%%%%%%%%%%%%%%%%%%
\begin{figure}[t!]
%\begin{minipage}{1.5\columnwidth}
	\centering
	\includegraphics[width=7cm,height=5cm]{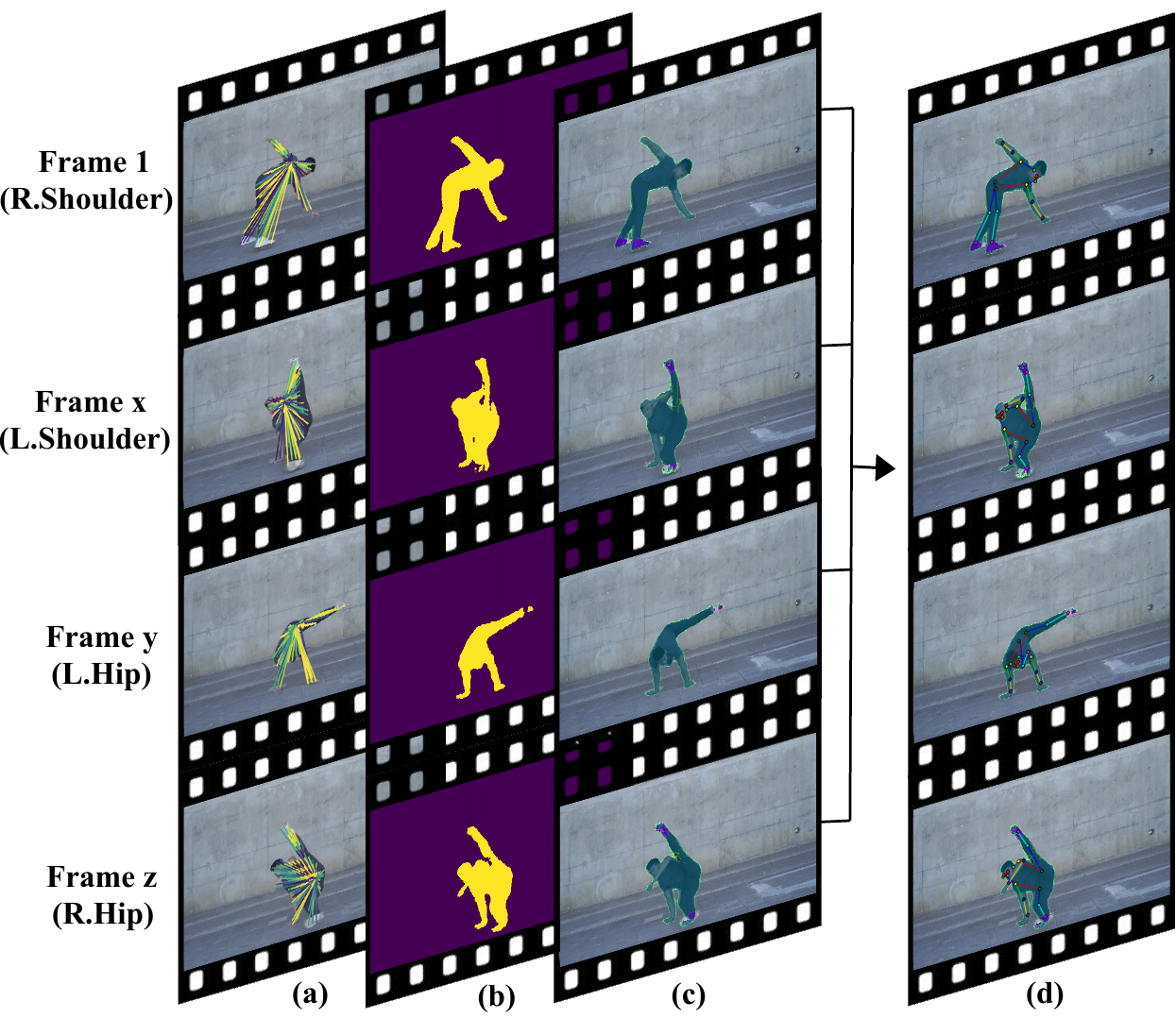}
	\caption{(a) Introduces MaskCentroid a dynamic high 
 confident keypoint; (b) presents a precise segmentation 
 map; (c) indicates instance-level segmentation; and 
 (d) shows unified representation of human pose and estimation.}
 %, i.e., human pose and instance segmentation.
        
	\label{fig:D_Mask_Centroid}
\end{figure}
%%%%%%%%%%%%%%%%%%%%%%%%%%%%%%%%%%%%%%%%%%%%%%%%%%%%%%%%%%%

\paragraph{Dynamic Center of Attraction}
%\label{subsec:IGO}
A significant innovation has been introduced in SegNet over the state-of-the-art \cite{ref86}, as shown in Figure \ref{fig:overview}. The previous model relied on a fixed centroid as a parameter to cluster mask pixels, which could lead to inferior results if the centroid is occluded in real-time scenarios. However, we allow the network to learn the optimal center of attraction by introducing the concept of a dynamic centroid. This is achieved by defining the high-confidence keypoint as a learnable parameter. 

This approach is especially valuable in scenarios where rapid occlusions occur during real-time operations, allowing the network to dynamically adjust the learned parameter and modify the center of attraction. As a result, the network can influence the location of the center of attraction by altering the embedding positions.
\begin{equation}
\phi_{j}(e_i) = \exp\left(-\dfrac{\| e_i - \left( \dfrac {1}{\vert N_j \vert} \sum_{e_{j} \in N_{j}} e_{j} \right) \|^2}{2{\sigma}^2_j}\right).
\end{equation}

In the inference phase, using keypoints as dynamic centroids for mask pixels effectively addresses complex scenarios where over 70\% of the human body is occluded. Our experimental study analyzes the effectiveness of both Static MaskCentroid \( SM_c \) and Dynamic MaskCentroid \( DM_c \) in human instance segmentation (\S{5.2}).

\paragraph{Instance-wise Gaussian Optimization}
\label{subsec:IGO}
To precisely align the predicted semantic maps, SegNet performs Gaussian smoothing \cite{ref84} at the instance level, \ie, \emph{instance-wise Gaussian optimization}. We apply instance-wise smoothing to reduce noise while retaining useful information, producing distinct semantic maps. The Gaussian kernel used for smoothing is defined as:

\begin{equation}
G(x,y) = \dfrac{1}{2\pi{\sigma}^2} \exp\left(-\dfrac{m_0^2 + m_1^2}{2{\sigma}^2}\right),
\end{equation}

where \( G(x,y) \) is the Gaussian kernel, \( \sigma \) is the standard deviation of the distribution, and \( (m_0, m_1) \) represents the pixel's coordinates within the kernel. We maintain \( \sigma \) within the range of 0.1 to 1. 

We find that a \( \sigma \) value close to 0.1 yields finer segmentation masks, particularly in scenarios where individuals are overlapped and entangled. Our ablation experiments support this observation and demonstrate the effectiveness of instance-wise smoothing (\S{5.3}).

% \paragraph{Instance-wise Gaussian Optimization}
% \label{subsec:IGO}
% %\red {
% To precisely align the predicted semantic maps, SegNet 
% performs Gaussian smoothing \cite{ref84} at the 
% instance-level, \ie, 
% \emph{instance-wise Gaussian optimization}. 
% We apply instance-wise smoothing to diminish noise and 
% retain useful information while producing distinct semantic 
% maps. 
% \begin{equation}
% G(x,y)= \dfrac{1}{2\pi{\sigma}^2} {exp}{-\dfrac{{m_0}^2+{m_1}^2}{2{\sigma}^2}},
% \end{equation}
% where $G(x,y)$ is the Gaussian kernel, $\sigma$ is the 
% standard deviation of the distribution, and $(m_0,m_1)$ 
% represents the pixel's coordinates within the kernel. 
% We maintain $\sigma$ to range from 0.1 to 1.%, ensuring precise instance segmentation aligns with the ground truth. 
% We find that a $\sigma$ value close to 0.1 yields finer 
% segmentation masks, particularly in scenarios where 
% individuals are overlapped and entangled.
% Our ablation experiments substantiate this observation and 
% the effectiveness of instance-wise smoothing  
% (\S{5.3}).
%}   

\subsection{PoseSeg Module}
\label{Pose_Seg_Module}

We introduce a new algorithm called PoseSeg, which simultaneously presents human pose estimation and instance segmentation, as illustrated in Fig. \ref{fig:D_Mask_Centroid}d. The PoseSeg module leverages high-level features generated by PoseNet and SegNet. Initially, keypoints and their coordinates are stored in a priority queue, facilitating the detection of body instances and the connection of adjacent keypoints. The pose kinematic graph is then followed to accurately estimate the human pose. Additionally, KDC performs instance-level segmentation by clustering pixels around centroids defined for each human instance. Specifically, pixels with a probability exceeding 0.5 are assigned to the corresponding human instances.

\section{Evaluation}
\label{sec:evaluation}
% We evaluate {\name} on three widely used benchmarks: COCO \cite{ref2}, CrowdPose \cite{ref92}, and OCHuman \cite{ref64}.
% The model is trained end-to-end using the COCO keypoint and segmentation training set. Ablations are conducted on the COCO \textit{val} set.
% The residual-based network ResNet-101 (RN-101) and ResNet-152 (RN-152) \cite{ref44} are used for training and testing. 
% The hyperparameters for training are: learning rate = $0.1\times e^{-4}$, image size $= 401 \times 401$, batch size $=4$, training epochs $ = 400$ and Adam optimizer. 
% We performed various transformations during model training, such as scale, flip, and rotate operations.
% Unless otherwise specified, a disk $D_R$'s radius is set to be $R = 32$.
We evaluate {\name} on COCO \cite{ref2}, CrowdPose \cite{ref92}, and OCHuman \cite{ref64} benchmarks.
The model is trained end-to-end using the COCO keypoint and segmentation training set, and ablations are conducted on the COCO \textit{val} set.
ResNet-101 (RN-101) and ResNet-152 (RN-152) \cite{ref44} are used for training and testing. 
Hyperparameters for training are: learning rate = $0.1\times e^{-4}$, image size $= 401 \times 401$, batch size $=4$, training epochs $ = 400$, and Adam optimizer is employed. 
Various transformations are applied during model training, such as scale, flip, and rotate operations.
Unless otherwise specified, a disk $D_R$'s radius is set to be $R = 32$.
\paragraph{Keypoint Results}
Table \ref{table:2} presents the performance of {\name} using the COCO keypoint \textit{test} set, outperforming the recent single-stage and top-down methods. 
% Point-Set \cite{ref75}, FCPose \cite{ref95}, DEKR \cite{ref77}, and CID \cite{ref96}. 
% {\name} is also better than state-of-the-art top-down techniques including recent RMPE \cite{ref36}, CPN \cite{ref35}, and HRNet \cite{ref97}. 
We also compare our method with bottom-up competitors. 
% HigherHRNet \cite{ref76} (multi-scale), SIMPLE \cite{ref79} (multi-scale), and Pose\textit{Plus}Seg \cite{ref86}. 
%Table \ref{table:2} shows the performance on COCO keypoint test dataset., respectively. It is cleared that the proposed {\name} outperforms bottom-up approaches, CMUPose \cite{ref32}, Associative Embedding \cite{ref3}, PersonLab \cite{ref52} and MultiPoseNet \cite{ref51}. 
{\name} with ResNet-152 yields an mAP of $76.1$, outperforming existing approaches by a large margin. Specifically, $5\%$ over Qu \textit{et al} \cite{ref99}, $4.9\%$ over DecentNet \cite{ref104}, $4.9\%$ over QueryPose \cite{ref114}, $3.3\%$ over Pose\textit{+}Seg \cite{ref91}, and $3.3\%$ over  GroupPose \cite{ref116}.
Table \ref{table:cowdpose} shows the results on the CrowdPose \textit{test} set compared to recent single-stage methods, top-down, and bottom-up models. 
{\name} (mAP 74.5) outperforms bottom-up OpenPose \cite{ref32}, HrHRNet \cite{ref76}, C.Atten. \cite{ref111}, and BUCTD \cite{ref112}. Table \ref{table:key_OCHuman} shows the results of {\name} compared with state-of-the-art models on OCHuman challenging dataset. 
We assess keypoint accuracy with top competitors LOGO-CAP \cite{ref97}, MIPNet \cite{ref97}, BUCTD \cite{ref112}, and CID \cite{ref97} both on \textit{val} and \textit{test} sets.

% {\name} improves 10.0\% over HGG \cite{ref74} (multi-scale) and 3.5\% over MIPNet \cite{ref78} using \textit{test} set.

%%%%%%%%%%%%%%%%%%%%%%%%%%%%%%%%%%%%%%%%%%%%%%%%%%%
%\vspace{15em}
\begin{table}[htb!]
\centering 
   \setlength{\tabcolsep}{0.3pt}
    \renewcommand{\arraystretch}{0.5}
 	\fontsize{8.5}{8.5}\selectfont	
	\begin{tabular}{ l|c| c c c c c }
	\hline
    Models & F.Work& AP & AP$^{50}$ & AP $^{75}$ & AP${^M}$ & AP${^L}$ \\
	\hline
	
	\textbf{Single-stage:}  \\
	% Point-Set \cite{ref75} & HR48  & 66.3& 87.7& 73.4& 64.9& 70.0 \\
    FCPose \cite{ref106} & RN101 &65.6& 87.9& 72.6& 62.1& 72.3  \\
    DEKR \cite{ref77} & HR32 &67.3& 87.9& 74.1& 61.5& 76.1 \\

    PETR\cite{ref103} &Swin-L& 70.5& 91.5& 78.7& 65.2& 78.0 \\
    
    CID \cite{ref107} & HR48 &70.7& 90.3& 77.9& 66.3& 77.8 \\

    ED-Pose\cite{ref115} &Swin-L& 72.2& 92.3& 80.9 & 67.6 & 80.0 \\

    RTMO\cite{ref117} &Darknet& 71.6& 91.1& 79.0 & 66.8 & 79.1 \\

    \hline
 
    \textbf{Top-down:}  \\
	Mask-RCNN \cite{ref34} & RN50 &63.1& 87.3& 68.7& 57.8& 71.4 \\
	Grmi\cite{ref33} & RN-101  &64.9& 85.5& 71.3& 62.3& 70.0 \\
	IntegralPose \cite{ref71} & RN-101 &67.8 &88.2 &74.8 &63.9 &74.0 \\
	% G-RMI$+$  & RN-101 &68.5 &87.1& 75.5& 65.8& 73.3& 73.3 \\
	CPN \cite{ref35} &RN-50 &72.1 & 91.4& 80.0& 68.7& 77.2 \\
	RMPE \cite{ref36} & PyraNet &72.3 & 89.2& 79.1& 68.0& 78.6 \\
	% CFN &- &72.6 & 86.1& 69.7& 78.3& 64.1& - \\
	
	% CPN (ensemble) \cite{ref35} &RN-Inc. &73.0 & 91.7& 80.9& 69.5& 78.1& 79.0 \\
	
	HRNet \cite{ref108}  &HR48 &75.5 & 92.5& 83.3& 71.9& 81.5 \\
	
	\hline
	
	\textbf {Bottom-up:}  \\
	OpenPose$\ast$\cite{ref32} & - & 61.8 & 84.9 & 67.5 & 57.1 & 68.2 \\
	Directpose$\ddagger$\cite{ref95} &RN-101 &64.8& 87.8& 71.1& 60.4& 71.5 \\
	% A.Em.$\ddagger$$\ast$\cite{ref3}& HG & 65.5& 86.8& 72.3& 60.6 & 72.6  \\
	PifPaf \cite{ref73} & RN-152 & 66.7 & - & - & 62.4 & 72.9 \\
	SPM \cite{ref96} &  HG & 66.9 & 88.5 & 72.9 & 62.6 & 73.1 \\
    PoseTrans\cite{ref100} &HrHR48& 67.4& 88.3& 73.9& 62.1& 75.1 \\

    SWAHR \cite{ref113} &HR32& 67.9& 88.9& 74.5& 62.4& 75.5 \\

	Per.Lab$\ddagger$\cite{ref52}&RN-152 &68.7& 89.0& 75.4& 64.1& 75.5 \\
	MPose\cite{ref51} & RN-101 & 69.6& 86.3& 76.6& 65.0& 76.3 \\
	% HGG $\ddagger$ & Hourglass  & 67.6& 85.1& 73.7& 62.7& 74.6& 71.3 \\

	HrHR$\ddagger$\cite{ref76} & HRNet  & 70.5& 89.3& 77.2& 66.6& 75.8 \\

    LOGP-CAP\cite{ref101} &HR48& 70.8& 89.7& 77.8& 66.7& 77.0 \\

     CIR\&QEM\cite{ref102} &HR48& 71.0& 90.2& 78.2& 66.2& 77.8 \\

	SIMPLE$\ddagger$\cite{ref79} & HR32 & 71.1& 90.2& 79.4& 69.1& 79.1 \\

     Qu \textit{et al} \cite{ref99} &HrHR48& 71.1& 90.4& 78.2& 66.9& 77.2 \\
     
    DecentNet\cite{ref104} & HR48 & 71.2& 89.0& 78.1& 66.7& 77.8 \\

    QueryPose\cite{ref114} &Swin-L& 72.2& 92.0& 78.8 & 67.3 & 79.4 \\

	Pose+Seg\cite{ref91} &RN-152& 72.8& 88.4& 78.7 & 67.8 & 79.4 \\

    GroupPose\cite{ref116} &Swin-L& 72.8& 92.5& 81.0 & 67.7 & 80.3 \\

	\hline
	\hline
	
	%\textit {\name} (ours): \\
	\textbf{{\name}} &\textbf{RN-101} & \colorbox{lightgray}{74.2}& \colorbox{lightgray}{89.0}& \colorbox{lightgray}{80.2}& \colorbox{lightgray}{69.3}& \colorbox{lightgray}{81.1} \\

	\textbf{{\name}} &\textbf{RN-152} &\colorbox{lightgray}{76.1}& \colorbox{lightgray}{92.9}& \colorbox{lightgray}{83.9}& \colorbox{lightgray}{71.1}& \colorbox{lightgray}{83.5} \\
    \hline

	\end{tabular}
	
	\caption{Performance comparison with recent works using \textbf{COCO} keypoint \textit {test} set. F.work indicates Framework, $+$ is trained on extra data, $\ast$ means refinement, $\ddagger$ is multi-scale results, HG indicates Hourglass network, and HR indicates High-Resolution Net.}  
	
	\label{table:2}

\end{table}
%\vspace{-1.9mm}
%%%%%%%%%%%%%%%%%%%%%%%%%%%%%%%%%%%%%%%%%%%%%%%%%%%%%%%%%%%%%%%%%
\begin{table}[htb!]
\centering 
    \setlength{\tabcolsep}{0pt}
    \renewcommand{\arraystretch}{0.5}
 	\fontsize{8.5}{8.5}\selectfont	
	\begin{tabular}{ l|c| c c c c c c} 
	\hline
    Models& F.Work & AP & AP$^{50}$ & AP$^{75}$& AP$_{E}$ & AP$_{M}$& AP$_{H}$ \\
	\hline
    \textbf{Single-stage:}  \\
    DEKR\cite{ref77} & HRNet & 65.7& 85.7& 70.4& 73.0& 66.4& 57.5\\
    PINet \cite{ref109} & HRNet & 68.9& 88.7& 74.7& 75.4& 69.6& 61.5 \\
    CID\cite{ref107} & HRNet & 72.3& 90.8& 77.9& 78.7& 73.0& 64.8 \\
    \hline
    \textbf{Top-down:}  \\
    MaskR\cite{ref34} & - & 57.2& 83.5& 60.3& 69.4& 57.9& 45.8 \\
    % Simp.Base. & - & 60.8& 81.4& 65.7& 71.4& 61.2& 51.2& 67.3 \\
    Al.Pose \cite{ref110} & - & 61.0& 81.3& 66.0 & 71.2& 61.4& 51.1 \\
	
    % Golda \textit{et al} & - & 65.5& -& -& 75.2& 66.6& 53.1& - \\
	% HrHRNet $\dagger$ & HRNet & 65.9& 86.4& 70.6& 73.3& 66.5& 57.9 &- \\
	J-SPPE \cite{ref92} & - & 66.0& 84.2& 71.5& 75.5& 66.3& 57.4 \\
    \hline
 \textbf {Bottom-up:}  \\
	OpenPose\cite{ref32} & - & -& -& -& 62.7& 48.7& 32.3  \\
	HrHR$\ddagger$\cite{ref76} & HRNet & 65.9& 86.4& 70.6& 73.3& 66.5& 57.9  \\
        C.Atten. \cite{ref111} & HRNet & 67.6& 87.7& 72.7& 75.8& 68.1& 58.9 \\

    BUCTD \cite{ref112} & HR48 & 72.9& -& -& 79.2& 73.4& 66.1  \\

	% DEKR $\dagger$ & HRNet & 67.3& 86.4& 72.2& 74.6& 68.1& 58.7 &- \\

	\hline
	\hline
	%\textit {\name} (ours):\\
	\textbf{\name}& \textbf{RN-101}  & \colorbox{lightgray}{71.6}& \colorbox{lightgray}{87.1}& \colorbox{lightgray}{75.2}& \colorbox{lightgray}{78.4}& \colorbox{lightgray}{71.9}& \colorbox{lightgray}{59.7}\\
	
	\textbf{\name}& \textbf{RN-152}  & \colorbox{lightgray}{74.5}& \colorbox{lightgray}{89.7}& \colorbox{lightgray}{76.8}& \colorbox{lightgray}{80.1}& \colorbox{lightgray}{74.8}& \colorbox{lightgray}{62.6}\\
	% \textbf{{\name} $\ddagger$}& ResNet101  & 0.723& 0.885& 0.774& 0.791& 0.717& 0.623& 0.788\\
	
	% \textbf{{\name} $\ddagger$}& ResNet152  & 0.741& 0.898& 0.787& 0.705& 0.724& 0.627& 0.802\\
	
	\hline
	\end{tabular}
    \caption{Performance comparison on \textbf{CrowdPose} keypoint %(human category)
    \textit{test} set. $\ddagger$ is multi-scale testing.}
	%\caption{Performance on COCO Segmentation (human category) \textit {val} set. The PersonLab \cite{ref52} and Pose2Seg \cite{ref64} results are obtain from their papers.}
	
	\label{table:cowdpose}
\end{table}
%\vspace{-15mm}
%%%%%%%%%%%%%%%%%%%%%%%%%%%%%%%%%%%%%%%%%%%%%%%%%%%%%%
% input test brings 2.6 mAP improvement compared to the single-scale test using the ResNet152 backbone feature extractor. Our best results on single-scale and multi-scale showed significant improvement over SOTA methods tested on CrowdPose dataset including OpenPose \cite{ref32}, HrHRNet \cite{ref76}, DEKR \cite{ref85}.\hfill \break
%}
%Specifically, the {\name} yields the best mAP of 0.728 on the ResNet-152 base architecture. 
%For comparison, bottom-up approaches that perform multi-scale inference and optimize their results using single-person pose estimation are also included in Table \ref{table:2}. 
%Some of the bottom-up approaches are also included in Table \ref{table:2} that performed multi-scale inference and optimized their results using a single-person pose estimation. 
%The test results also show that the performance of {\name} surpasses that of top-down approaches, i.e., Mask-RCNN \cite{ref34}, G-RMI \cite{ref33}, Integral Pose Regression \cite{ref71}, and CPN \cite{ref35}.
%
%%%%%%%%%%%%%%%%%%%%%%%%%%%%%%%%%%%%%%%%%%%%%%%%%%%%%%%%%%%%
\begin{table}[htb!]
\centering 
     \setlength{\tabcolsep}{5pt}
    \renewcommand{\arraystretch}{0.3}
    \fontsize{8.5}{8.5}\selectfont	
	\centering
 	% \fontsize{8.5}{8.5}\selectfont	
	\begin{tabular}{ l|c| c c }
	\hline
    Models& F.Work & Val mAP & Test mAP \\
	\hline
	
	HGG \cite{ref74} & HG &35.6& 34.8  \\

        DEKR \cite{ref77} &HRNet  & 37.9& 36.5 \\

        HrHR \cite{ref76} &HrHR32  & 40.0& 39.4 \\

        LOGO-CAP \cite{ref101} &HR48  & 41.2& 40.4 \\
        
        %HGG $\ddagger$ \cite{ref74}& HG &41.8& 36.0  \\

	MIPNet \cite{ref97} &RN-101  & 42.0& 42.5 \\

        BUCTD \cite{ref112} &HrHR32  & 44.1& 43.5 \\

        CID \cite{ref107} &HR32  & 45.7& 44.6 \\

	\hline
	\hline
	%\textit {\name} (ours):\\
	\textbf{\name}& \textbf{RN-101}  & \colorbox{lightgray}{44.1}& \colorbox{lightgray}{44.6} \\
	
	\textbf{\name}& \textbf{RN-152}  & \colorbox{lightgray}{46.3}& \colorbox{lightgray}{46.0} \\
	\hline
	\end{tabular}

    \caption{Performance using \textbf{OCHuman} keypoint \textit{val} and \textit{test} datasets.}
	%\caption{Performance on COCO Segmentation (human category) \textit {val} set. The PersonLab \cite{ref52} and Pose2Seg \cite{ref64} results are obtain from their papers.}
	
	\label{table:key_OCHuman}
	%\vspace{-15mm}
\end{table}

\begin{table}[htb!]
\centering 
    \setlength{\tabcolsep}{0.3pt}
    \renewcommand{\arraystretch}{0.3}
    \fontsize{8.5}{8.5}\selectfont	
  %   \renewcommand{\arraystretch}{0.6}
 	% \fontsize{8.5}{8.5}\selectfont	
	\begin{tabular}{ l|c| c c c c c } 
	\hline
    Models& F.Work & AP & AP$^{50}$ & AP$^{75}$ & AP${^M}$ & AP${^L}$ \\
	\hline
	
	MaskRCNN\cite{ref34} & RN-101 & 37.1& 60.0& 39.4& 39.9& 53.5 \\
	Per.Lab$\dagger$\cite{ref52} & RN-101& 37.7& 65.9& 39.4& 48.0& 59.5 \\
	Per.Lab$\dagger$\cite{ref52} &RN-152 & 38.5& 66.8& 40.4& 48.8& 60.2 \\
	Per.Lab$\ddagger$\cite{ref52} &RN-101 &41.1& 68.6& 44.5& 49.6& 62.6 \\
	Per.Lab$\ddagger$\cite{ref52} &RN-152 & 41.7& 69.1& 45.3& 50.2& 63.0 \\
	Pose\textit{+}Seg\cite{ref91} & RN-152  & 44.5& 79.4& 47.1& 52.4& 65.1 \\

	\hline
	\hline
	%\textit {\name} (ours):\\
	\textbf{{\name}}& \textbf{RN-101}  & \colorbox{lightgray}{45.7}& \colorbox{lightgray}{80.4}& \colorbox{lightgray}{47.8}& \colorbox{lightgray}{53.5}& \colorbox{lightgray}{67.4} \\
	
	\textbf{{\name}}& \textbf{RN-152}  & \colorbox{lightgray}{47.6}& \colorbox{lightgray}{81.8}& \colorbox{lightgray}{48.7}& \colorbox{lightgray}{54.6}& \colorbox{lightgray}{67.8} \\
	\hline
	\end{tabular}
	\caption{Performance comparison on \textbf{COCO} Segmentation %(human category)
	\textit{test} set. $\dagger$ is single-scale  testing. $\ddagger$ is multi-scale testing.}
	%\caption{Performance on COCO Segmentation (human category) \textit {test} set. The PersonLab \cite{ref52} and Pose2Seg \cite{ref64} results are obtain from their papers.}	
	\label{table:4}
	%\vspace{-15mm}
\end{table}

%%%%%%%%%%%%%%%%%%%%%%%%%%%%%%%%%%%%%%%%%%%
\begin{table}[htb!]
    \setlength{\tabcolsep}{4pt}
    \renewcommand{\arraystretch}{0.5}
    \fontsize{8.5}{8.5}\selectfont	
	\label{}
	\centering
 	%\fontsize{8.5}{8.5}\selectfont	
	\begin{tabular}{ l|c| c c } 
	\hline
    Models& F.Work & Val mAP & Test mAP \\
	\hline
	
	Pose2Seg \cite{ref64} &RN-50-fpn  & 54.4& 55.2 \\

	\hline
	\hline
	%\textit {\name} (ours):\\
	\textbf{\name}& \textbf{RN-101}  & \colorbox{lightgray}{56.7}& \colorbox{lightgray}{57.0} \\
	
	\textbf{\name}& \textbf{RN-152}  & \colorbox{lightgray}{58.3}& \colorbox{lightgray}{59.6}\\
	\hline
	\end{tabular}

    \caption{Performance comparison using \textbf{OCHuman} segmentation \textit{val} and \textit{test} datasets.}
	%\caption{Performance on COCO Segmentation (human category) \textit {val} set. The PersonLab \cite{ref52} and Pose2Seg \cite{ref64} results are obtain from their papers.}
	\label{table:Seg_OCHuman}
	%\vspace{-10mm}
\end{table}
%%%%%%%%%%%%%%%%%%%%%%%%%%%%%%
%Figure 5 Place
%%%%%%%%%%%%%%%%%%
\paragraph{Segmentation Results}
Table \ref{table:4} presents instance-level segmentation results using COCO segmentation \textit{test} sets. {\name} delivered a top accuracy of 47.6 mAP and improved the AP by 10.5\% over Mask-RCNN \cite{ref34}, 5.9\% over PersonLab \cite{ref52} (multi-scale), and 3.1\% over Pose\textit{+}Seg \cite{ref91}. Table \ref{table:Seg_OCHuman} presents the segmentation performance on the OCHuman \textit{val} and \textit{test} sets. {\name} (mAP 58.3), demonstrating a significant improvement of 3.9\% and 4.4\% over Pose2Seg \cite{ref64} on the \textit{val} and \textit{test} sets, respectively.

\begin{figure}[t!]
%\begin{minipage}{1.5\columnwidth}
	\centering
        \includegraphics[width=8cm,height=10cm]{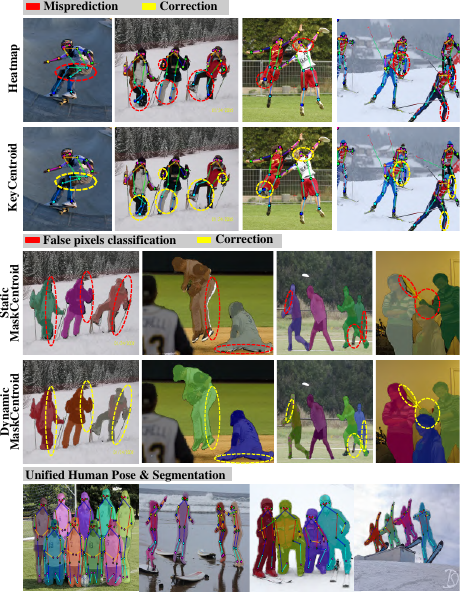}

 \caption{Visual results from various components of the system reveal initial mispredictions and inaccuracies in the keypoint heatmap (first row), corrected by KeyCentroid (second row). False pixel classification in segmentation with Static MaskCentroid (third row) was resolved using Dynamic MaskCentroid (fourth row). Unified human pose and segmentation are shown in the fifth row.}
 %\caption{Visual results from different components of the system. We observed mispredictions and inaccurate connections in the keypoint heatmap (first row), which were rectified with the implementation of KeyCentroid (second row). Furthermore, we pointed out the false pixels classification in the segmentation task using Static MaskCentroid (third row), yet this was resolved through the use of Dynamic MaskCentroid (fourth row). Visuals of the unified human pose and segmentation (fifth and sixth row).}
 %, i.e., human pose and instance segmentation.        
	\label{fig:visualization}
\end{figure}

\begin{table}[htb!]
    \setlength{\tabcolsep}{3pt}
    \renewcommand{\arraystretch}{0.3}
    \fontsize{8.5}{8.5}\selectfont	
	\label{}
	\centering
 	%\fontsize{8.5}{8.5}\selectfont	
	\begin{tabular}{ l|c| c c c } 
	\hline
    Models& F.Work & Val mAP & Test mAP \\
	\hline
	
	CRMH \cite{ref93} &-& 32.9& 33.9 \\
        ROMP \cite{ref94} &RN-50  & 55.6& 54.1 \\
        ROMP+CAR \cite{ref94} &RN-50  & 58.6& 59.7 \\
	\hline
	\hline
	%\textit {\name} (ours):\\
	\textbf{\name}& \textbf{RN-101}  & \colorbox{lightgray}{86.3}& \colorbox{lightgray}{87.1} \\
	
	\textbf{\name}& \textbf{RN-152}  & \colorbox{lightgray}{88.1}& \colorbox{lightgray}{89.7}\\
	\hline
	\end{tabular}

    \caption{Comparisons with 3D methods on the \textbf{CrowdPose} benchmark using $AP^{.50}$ evaluation metric.}
	%\caption{Performance on COCO Segmentation (human category) \textit {val} set. The PersonLab \cite{ref52} and Pose2Seg \cite{ref64} results are obtain from their papers.}
	\label{table:3d_pose}
	%\vspace{-10mm}
\end{table}
\begin{figure}[htb!]
%\begin{minipage}{1.5\columnwidth}
	\centering
        \includegraphics[width=6.6cm,height=5.0cm]{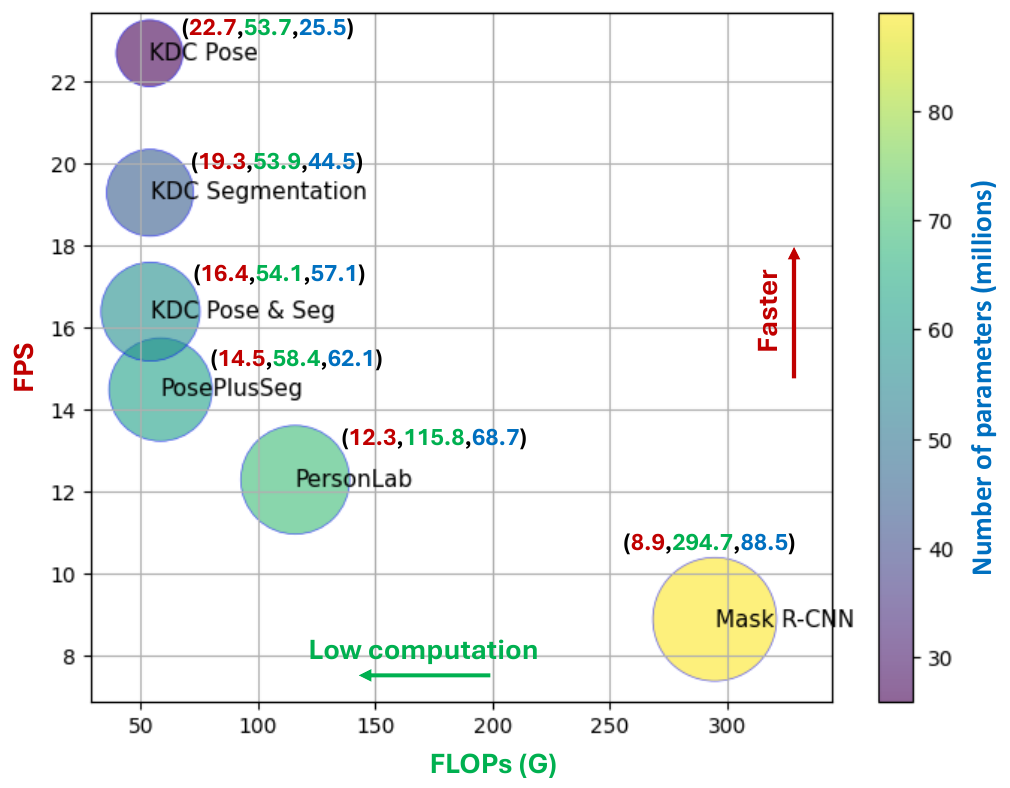}

 \caption{Computational cost with the representative sister models. Models are tested on a single Titan RTX.}
 %, i.e., human pose and instance segmentation.        
	\label{fig:12}
\end{figure}
%%%%%%%%%%%%%%%%%%%%%%%%%%%%%%%%%%%%%%%%%%%%%%%%%%%%
% %%%%%%%%%%%%%%%%%%%%%%%%%%%%%%%%%%%%%%%%%%%%%%

\paragraph{Comparing 2D  vs.\ 3D Pose Estimation}
We also compare the pose performance with state-of-the-art 3D models CRMH \cite{ref93} and ROMP \cite{ref94} in crowded scenes. We calculate the average precision ($AP^{0.5}$) between the 2D projection of the 3D pose on the Crowdpose  \textit{val} and \textit{test} sets shown in Table \ref{table:3d_pose}.

\paragraph{Computational Cost}
%\red{ 
%Testing the system on various sizes of images are an import part of the testing phase. 
We calculate the computational cost and FPS using an image resolution of $401 \times 401$. Fig. \ref{fig:12} shows that {\name} has fewer parameters, high FPS, and lower computational complexity compared to the representative models Mask R-CNN \cite{ref34}, PersonLab \cite{ref52}, and Pose\textit{+}Seg \cite{ref91}.

% and highest mAP compared with Hourglass \cite{ref22} and CPN \cite{ref35}.
% %Table \ref{table:11} shows that the {\name} has a lowest FLOPs and highest mAP by using ResNet-50 compare to Hourglass \cite{ref22} on same input size. 
% %{\name} has lead on CPN \cite{ref35} using the same settings. 
% {\name} with ResNet-101 and ResNet-152 also incurs lower FLOPs and number of parameters compared with the representative top competitors, including SimpleBaseline \cite{ref70}, HRNet \cite{ref72}, DEKR \cite{ref77}, PersonLab \cite{ref52}, and Pose\textit{Plus}Seg \cite{ref86}. Table \ref{table:12} presents the inference time and runtime measurements of {\name} on a single GPU (Titan RTX). 

%Results on ResNet-101 and ResNet-152 are satisfactory compared to SimpleBaseline \cite{ref70}, HRNet \cite{ref72}, and DEKR \cite{ref77} by using less amount of FLOPs and number of parameters by 256 $\times$ 192 input image size. 

% We also measure system inference time compared with SOTA multi-task models like Mask-RCNN \cite{ref34}, Pose2seg \cite{ref64}, MultiPoseNet \cite{ref51}, and Pose\textit{Plus}Seg \cite{ref86}. 

%demonstrating its efficiency in processing for human pose, instance segmentation, and combine human pose and segmentation.

%We perform execution test on 2 different GPUs for better run time assessment as illustrated in 
%}

\section{Ablation Experiments}
\label{sec:ablation}

% %\red{
% In this section we discuss ablation experiments to qualitatively analyze each component of the system. 
% Our experiments focus on the impact of SKHM with KeyCentroid (\S\ref{subsec:Imp_SKHM}), BHM (\S\ref{subsec:Imp_BHM}), MaskCentroid (\S\ref{subsec:Ab_Maskoffset}), point-wise and instance-wise Gaussian optimization (\S\ref{subsec:Gaussian_opt}).
%,  Gaussian optimization (\S\ref{subsec:Instance-wise_Gaussian}). 
%This way, we validate the effectiveness of our system from various aspects. 
% Unless otherwise specified, all the experiments are evaluated using the COCO \emph{val} dataset.
%}

\subsection{KHDR and KeyCentroid}
\label{subsec:Imp_SKHM}
%
%we run a set of ablation experiments to qualitatively analyze the impact of each component of {\name}.
 %\red{
% Initially, we compared the keypoint heatmap using disk representation (KHDR) and keypoint detection algorithms dependent on Gaussian peak-based detection.
% %We first validate Strong Keypoint Heat Map (SKHM) with a number of keypoint detection algorithms that depends on keypoint heatmaps for keypoint detection and pose estimation. 
% Table \ref{table:5} presents the performance of KHDR and KeyCentroid $k_{c}$ with state-of-the-art bottom-up approaches including CMU-Pose \cite{ref32}, MultiPoseNet \cite{ref51}, PersonLab \cite{ref52}, HGG \cite{ref74}, SimpleBaseline \cite{ref70}, and Pose\textit{Plus}Seg \cite{ref86}. KDC's KHDR outperforms the state-of-the-art methods, yielding an mAP of 74.8 using ResNet152.
%  %In the ablation study it worth noticing that the keypoint predicted by SKHM shows high average precision (AP) rate in all categories. 
%  %Using the ResNet152, the SKHM shows a significant improvement and yields an mAP of 0.724 in total.
%  In addition, we defined KeyCentroids and aggregated them with the KHDR to find the optimal 2D keypoint coordinates. 
%  Thus, {\name} further improved the keypoint accuracy by 2.7 points and secured a 77.5 mAP resulting in a significant improvement over Pose\textit{Plus}Seg \cite{ref86}. Figure \ref{fig:visualization} shows a visual display of keypoint heatmap improvement using KeyCentroid. 
 %We calculate the keypoint confidence score using the keypoint disk.

Initially, we evaluate the performance of the proposed KHDR and examine its effectiveness with and without the integration of $K_c$, as presented in Table \ref{table:5}. Through our ablation study, we observe that the combination of KHDR and $K_c$ is a highly effective approach for human pose estimation, particularly in challenging scenarios and dynamic movement of the human body. Fig. \ref{fig:visualization} shows the visual performance of keypoint heatmap improved by KeyCenroid. 
Fig. \ref{fig:7} shows the predicted confidence score of 17 keypoints using the keypoint disk at radius $R=8$, $16$, and $32$. 

\begin{table}[htb!]

\centering 
    \setlength{\tabcolsep}{7pt}
    \renewcommand{\arraystretch}{0.5}
    \fontsize{8.5}{8.5}\selectfont	
	\begin{tabular}{ l|c c c c c}
	\hline
    \textbf{{\name}} w and w/o\\ 
    \textbf{KHDR \kern 3 em $k_{c}$}& \textbf{AP} & \textbf{AP}$^{.50}$ & \textbf{AP} $^{.75}$ & \textbf{AP}${^M}$ & \textbf{AP}${^L}$ \\
	\hline
	% CMU-Pose & 61.0& 84.9& 67.5& 56.3& 69.3\\
	% MultiPoseNet & 64.3& 88.2& 75.0& 59.6& 73.9  \\
	% PersonLab  & 66.5& 86.2& 71.9& 62.3& 73.2\\
	% HGG  & 68.3& 86.7 & 75.8& -& -\\
	% SimpleBaseline  & 72.0& 89.3& 79.8& 68.7& 78.9\\
	% Pose\textit{Plus}Seg & 74.4& 89.4& 74.8& 67.5& 81.1\\

	% \hline
	\hline

	\textbf{\kern 1 em \checkmark} & 74.8& 89.7& 75.6 & 70.3 & 79.1\\
	\textbf{ \kern 5.9 em \checkmark}  &76.2& 91.8 & 78.9& 72.5& 82.7\\
	\textbf{\kern 1 em \checkmark \kern 4.3 em \checkmark}  &77.5& 94.9& 86.4& 73.8& 84.6\\

	\hline
	\end{tabular}

	\caption{Performance of KHDR with and without KeyCentroid $k_{c}$ mechanism.}
	\label{table:5}
	%\caption{Keypoint detection comparison between Strong Keypoints Heat Map(SKHM) and keypoint heatmap. Results of CMU-Pose \cite{ref32}, MultiPoseNet \cite{ref51} and PersonLab \cite{ref52} are referred from their paper.}
    %\vspace{-10mm}
\end{table}

%%%%%%%%%%%%%%%%%%%%%%%%%%%%%%%%%%%%%%%%%%%%%%%

%%%%%%%%%%%%%%%%%%%%%%%%%%%%%%%%%%%%%%%%%%%%%%%%%%%%%%%%%%%
\begin{figure}[htb!]
%\begin{minipage}{1.5\columnwidth}
	\centering
        \includegraphics[width=8.3cm,height=4.2cm]{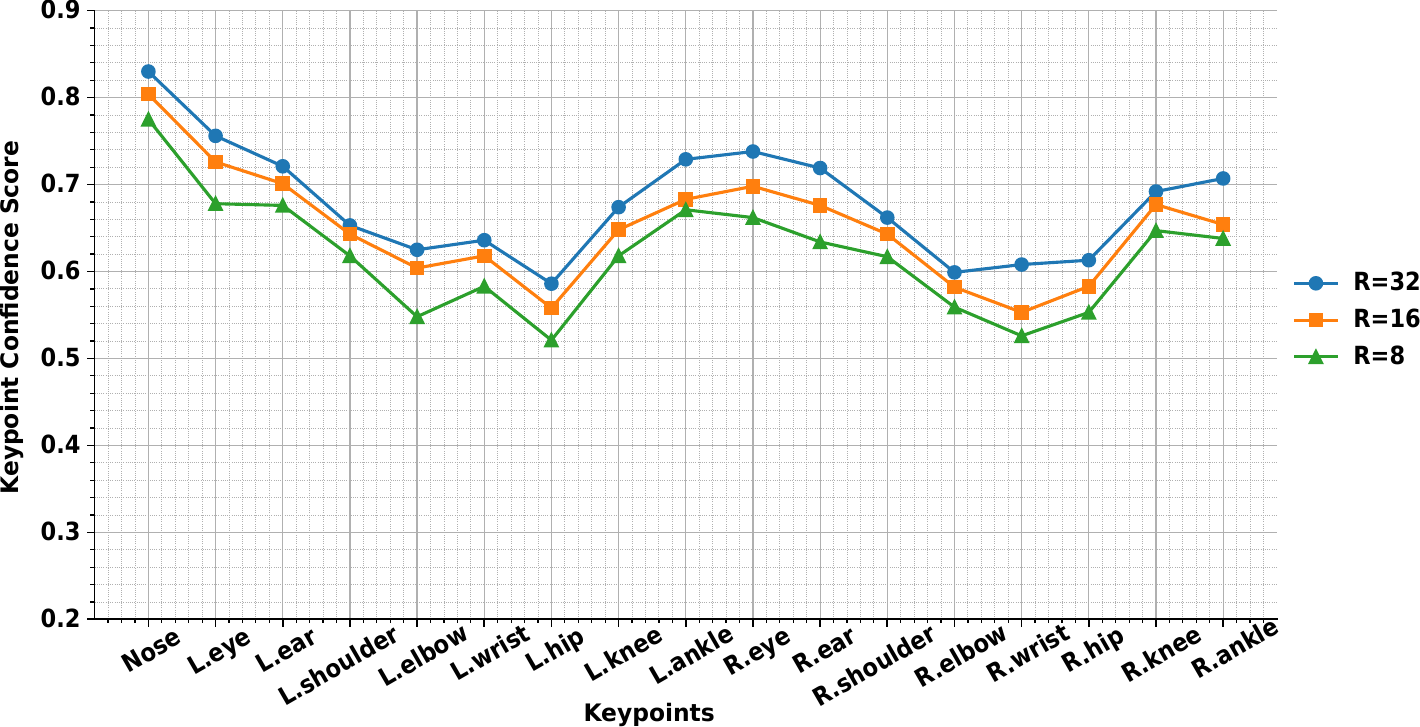}

 \caption{Left(L) and right(R) keypoint confidence score with 
 varied disk radius $R=\{32, 16, 8\}$.}
 %, i.e., human pose and instance segmentation.        
	\label{fig:7}
\end{figure}

%%%%%%%%%%%%%%%%%%%%%%%%%%%%%

\begin{figure}[htb!]
%\begin{minipage}{1.5\columnwidth}
	\centering
	\includegraphics[width=8.5cm,height=4.6cm]{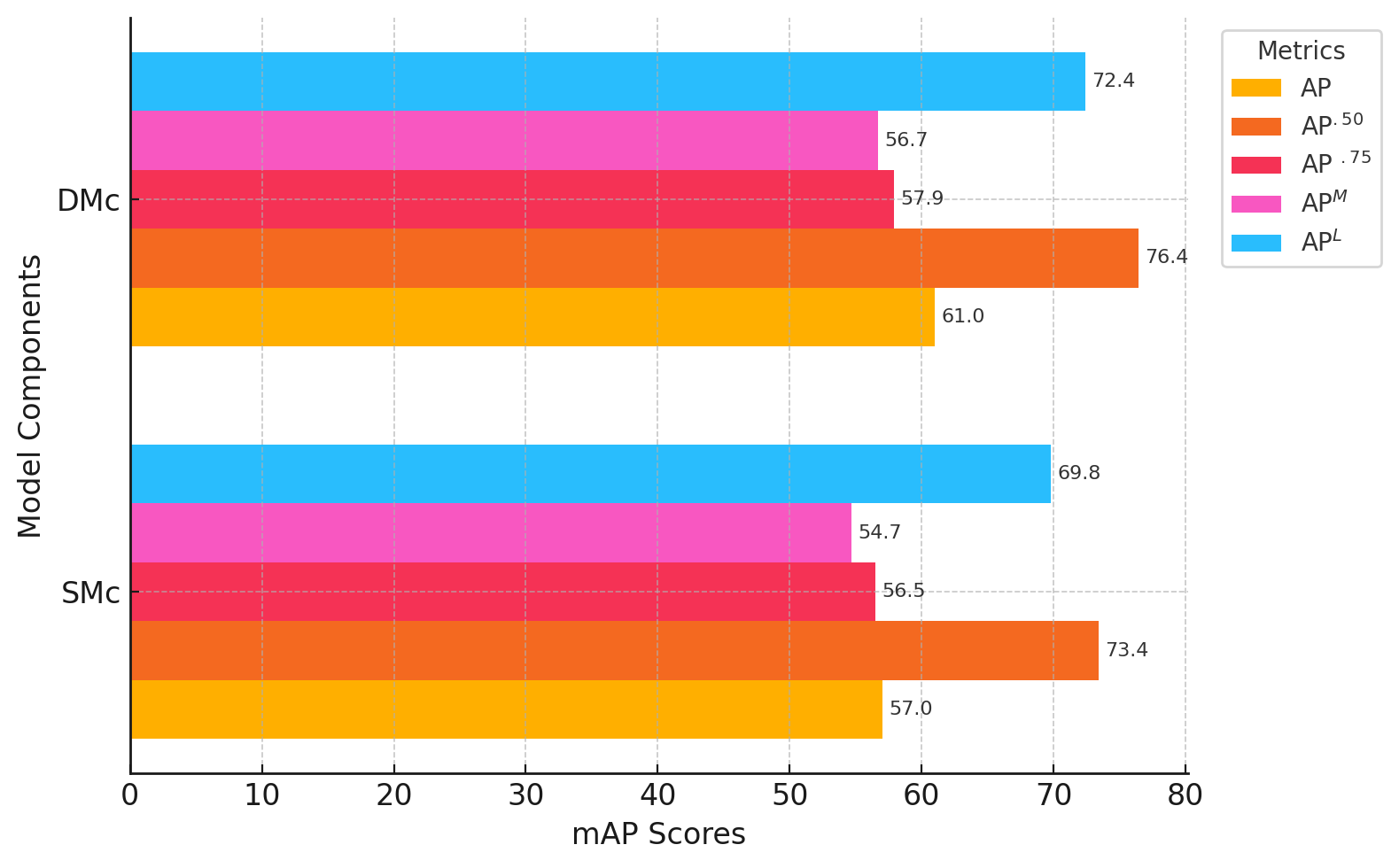}
\caption{Performance of $SM_c$ and $DM_c$ on human instance-level 
segmentation.} 
 \label{fig:SMc_DMc}
\end{figure}

%%%%%%%%%%%%%%%%%%%%%%%%%%%%%%%%%%%%%%%%%%%%%%%%%%%%
\begin{figure}[hbt!]
    \begin{minipage}{0.48\columnwidth}
    \centering
	%\captionsetup[subfigure]{justification=centering}
	\includegraphics[width=\columnwidth, height=3.1cm]{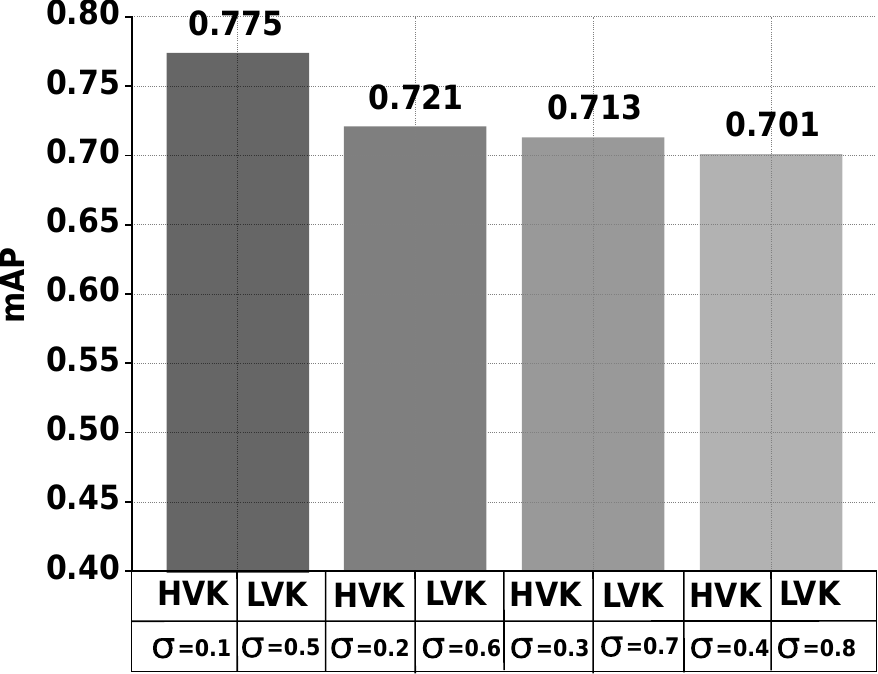}
  \caption{Point-wise Gaussian optimization with different $\sigma$ values for HVK and LVK.}
	%\caption{Keypoint detection results with varied $\sigma$ values.}
 	\label{table:Gaussain}
	 \end{minipage}
	 \hfill
	 \begin{minipage}{0.48\columnwidth}
	\centering
	%\captionsetup[subfigure]{justification=centering}
		\includegraphics[width=\columnwidth, height=3.1cm]{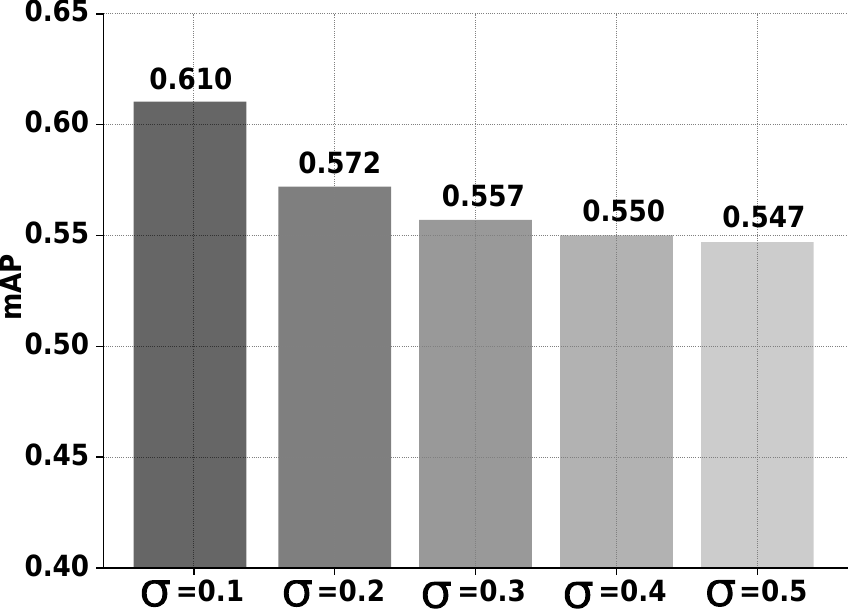}
  \caption{Instance-wise Gaussian optimization with different $\sigma$. Small $\sigma$ provides precise instance mask.}
  %\caption{Segmentation results with varied $\sigma$ values.}
 	\label{table:Instance-wise_Gaussian}
    \end{minipage}
	 
\end{figure}

\subsection{Static vs.\ Dynamic MaskCentroids}
\label{subsec:Ab_Maskoffset}

We analyze the Static MaskCentroid ($SM_c$) and the Dynamic 
MaskCentroid ($DM_c$), with the results presented in 
Fig.~\ref{fig:SMc_DMc}. The exceptional performance of the proposed $DM_c$ approach demonstrates its effectiveness in human body segmentation, particularly in scenarios involving dynamic human body movement. Fig.~\ref{fig:visualization} shows the visual performance of $SM_c$ improved by $DM_c$. This capability significantly contributes to advancements in instance-level segmentation.

\subsection{Point \& Instance-wise Gaussian Optimization}
\label{subsec:Gaussian_opt}
%\paragraph{Point-wise Gaussian Optimization}
We generated keypoint heatmap utilizing point-wise Gaussian optimization using $ 0.1 \le \sigma \le1$. Fig. \ref{table:Gaussain} summarizes the mAP for different $\sigma$ with high variation of keypoints (e.g., wrist, ankle, elbow, and knee) and low variation of keypoints (e.g., nose, shoulder, hip). %The high accuracy is obtained by using the $\sigma = 0.1$ for HVK and $\sigma = 0.5$ for LVK. 
 %\vspace{-4mm}
%%%%%%%%%%%%%%%%%%%%%%%%%%%%%%%%%%%%%%%%%%%%%%%%%%%%%%%%%%%%%%%%%
% \begin{table}[hbt!]
    
%     \setlength{\tabcolsep}{4pt}
%     \renewcommand{\arraystretch}{0.9}
	
% 	\label{}
% 	%\vspace{10mm}
	
% 	\centering
%  	\fontsize{8}{8}\selectfont	
% 	\begin{tabular}{ c|c c} 
%     High variation keypoints & Low variation keypoints & mAP  \\

% 	%\hline
% 	\hline
% 	{$\sigma$} = 0.1  & {$\sigma$} = 0.5 & 0.724  \\
% 	{$\sigma$} = 0.2  & {$\sigma$} = 0.6 & 0.720  \\
% 	{$\sigma$} = 0.3  & {$\sigma$} = 0.7 & 0.713  \\
% 	{$\sigma$} = 0.4  & {$\sigma$} = 0.8 & 0.701  \\
% 	\end{tabular}

% 	\caption{Keypoint detection results with different $\sigma$ values.}
% 	\label{table:Gaussain}
%     \vspace{-4mm}
% \end{table}

%\begin{figure}[h!]
%	\centering{\includegraphics[width=0.66\columnwidth]{figures/graph_1.pdf}}
%	%\caption{MaskCentroids where the right shoulder is the center of attraction.}
%	\caption{Keypoint detection results with different $\sigma$ values.}
% 	\label{table:Gaussain}
%   %\vspace{-4mm}
%\end{figure}
%%%%%%%%%%%%%%%%%%%%%%%%%%%%%%%%%%%%%%%%%%%%%%%%%%%%%%%5

%\subsection{Impact of Instance-wise Gaussian Optimization}
%\label{subsec:Instance-wise_Gaussian}
Finally, we examine the impact of instance-wise Gaussian optimization 
on the instance segmentation task. 
We tested the sensitivity of $\sigma$ ranging from $0.1$ to $0.5$ 
on human instance segmentation. 
Fig.~\ref{table:Instance-wise_Gaussian} shows the results with 
different $\sigma$ values, where low $\sigma$ provides precise 
segmentation mask and performs better in crowded cases.

\section{Conclusion}
\label{sec:outro}
This paper considers the challenge of unified human pose estimation 
and instance-level segmentation, particularly in complex multi-person 
dynamic movement scenarios. {\name} generate keypoint heatmaps by 
defining keypoint disks and KeyCentroid to determine the optimal 2D 
keypoint coordinates within the specified keypoint disk. 
Additionally, MaskCentroid is introduced, representing highly 
confident keypoint as dynamic centroids to cluster the mask pixels 
with the correct instance in the embedding space, even under 
significant occlusion or body movement. The effectiveness of {\name} 
is evaluated on COCO, CrowdPose, and OCHuman benchmarks and proves to 
be a highly effective approach for unified human pose estimation and 
instance-level segmentation.

\section*{Acknowledgments}

This work was partly funded by the Natural Sciences and Engineering Research Council of Canada (NSERC) grant RGPIN-2021-04244, the Institute of Information and Communications Technology Planning and Evaluation (IITP) grant IITP-2025-RS-2020-II201741, RS-2022-00155885, RS-2024-00423071 funded by the Korea government (MSIT).

\bibliographystyle{named}
\bibliography{ijcai25}

\end{document}